\documentclass[letterpaper, 10 pt, conference]{ieeeconf}  

\IEEEoverridecommandlockouts                              

\usepackage{graphics} 
\usepackage{epsfig} 
\usepackage{mathptmx} 
\usepackage{times} 
\usepackage{amsmath} 
\usepackage{amssymb}  
\usepackage{booktabs}   
\usepackage{cite}
\usepackage{algorithmic}
\usepackage{algorithm}
\usepackage{multirow}
\usepackage{graphicx} 

\usepackage{float} 
\usepackage{bm}

\usepackage{marvosym}

\usepackage[caption=false,font=footnotesize,labelfont=rm,textfont=rm]{subfig}
\DeclareMathAlphabet{\mathcal}{OMS}{cmsy}{m}{n}
\DeclareSymbolFont{largesymbols}{OMX}{cmex}{m}{n}

\title{\LARGE \bf
Fast and Communication-Efficient Multi-UAV Exploration Via Voronoi Partition on Dynamic Topological Graph 
}

\author{Qianli Dong\textsuperscript{\Cross}, Haobo Xi\textsuperscript{\Cross}, Shiyong Zhang$^{\star}$, Qingchen Bi, Tianyi Li, Ziyu Wang, and Xuebo Zhang
\thanks{This work was sponsored by Natural Science Foundation of China under grant 62293513/62293510 and 62303249. All authors are with the College of Artificial Intelligence, Nankai University, Tianjin, China. {\tt\small \{qianlidong,xihaobo\}@mail.nankai.edu.cn, \tt\small zhangshiyong@nankai.edu.cn}}
\thanks{{\Cross} These authors contributed equally to this work.}
\thanks{${\star}$ Corresponding Author}%
}

\begin{document}

\maketitle
\thispagestyle{empty}
\pagestyle{empty}

\begin{abstract}

Efficient data transmission and reasonable task allocation are important to improve multi-robot exploration efficiency. However, most communication data types typically contain redundant information and thus require massive communication volume. Moreover, exploration-oriented task allocation is far from trivial and becomes even more challenging for resource-limited unmanned aerial vehicles (UAVs). In this paper, we propose a fast and communication-efficient multi-UAV exploration method for exploring large environments. We first design a multi-robot dynamic topological graph (MR-DTG) consisting of nodes representing the explored and exploring regions and edges connecting nodes. Supported by MR-DTG, our method achieves efficient communication by only transferring the necessary information required by exploration planning. To further improve the exploration efficiency, a hierarchical multi-UAV exploration method is devised using MR-DTG. Specifically, the \emph{graph Voronoi partition} is used to allocate MR-DTG's nodes to the closest UAVs, considering the actual motion cost, thus achieving reasonable task allocation. To our knowledge, this is the first work to address multi-UAV exploration using \emph{graph Voronoi partition}. The proposed method is compared with a state-of-the-art method in simulations. The results show that the proposed method is able to reduce the exploration time and communication volume by up to 38.3\% and 95.5\%, respectively. Finally, the effectiveness of our method is validated in the real-world experiment with 6 UAVs. We will release the source code to benefit the community.

\end{abstract}

\section{INTRODUCTION}


Autonomous exploration using unmanned aerial vehicles (UAVs) has been extensively investigated \cite{rh_nbvp, fuel, fast_sample} and increasingly applied in many practical applications, such as search-and-rescue \cite{zhang2022fast} and structural inspection \cite{quin2017experimental}. However, it is intractable for a single UAV to explore large environments. Therefore, multi-UAV systems have been employed to further improve the exploration efficiency \cite{racer, ethz, tolstaya2021multi}. Up to now, efficient communication and reasonable task allocation are still challenging for multi-UAV exploration.

Since the map-sharing process requires the most communication resources, existing methods take a lot of effort to optimize this process for multi-robot exploration. To coordinate and synchronize exploration information, the work in \cite{yu2021smmr} directly shares point cloud information organized via submaps at a low frequency, which might lead to vulnerable coordination. The volumetric map is a sparser representation of the environment than point clouds. In the work \cite{racer}, newly observed voxels are grouped into chunks for incremental and immediate communication. While this method can efficiently reduce the communication volume, transferring chunks is still communication-consuming, especially as the number of employed UAVs increases. Different from the above methods, topological maps are used in the work \cite{MR-TopoMap} for multi-robot exploration, which can significantly reduce the data transmission burden. However, the visibility requirements between vertices of the topological map make it hard to be used in cluttered environments.  

\begin{figure}[tp]
	\centerline{\includegraphics[width=0.45\textwidth]{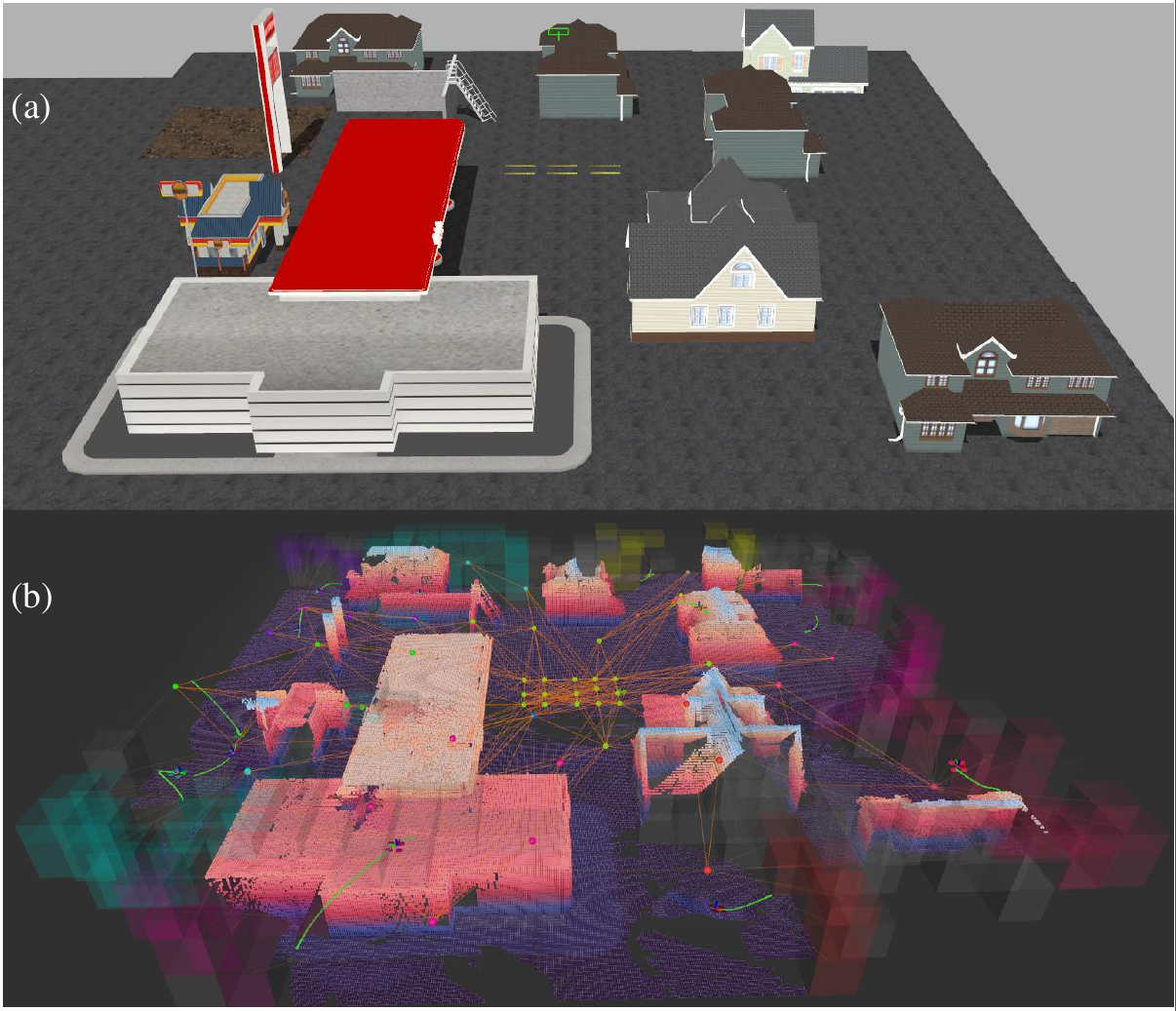}} 
	\caption{An exploration instance of the proposed multi-UAV exploration, where 15 UAVs are exploring the environment with MR-DTG.
		(a) The simulation environment of a city scene.
		(b) Exploration results using the proposed method.
		Dots represent history nodes, and cubes represent explorable regions.
		The colors of history nodes correspond to the UAVs they are assigned to during the global partition, and the colors of explorable regions correspond to the UAVs they are assigned to during the local partition.
		Video of the experiments is available at: https://www.youtube.com/watch?v=AtG9stNVjX0\&t=1s.
	}
	\label{example}
\end{figure}

On the other hand, a reasonable task allocation strategy is also crucial for improving the efficiency of multi-robot exploration. Previous methods usually coordinate multi-robot exploration with a centralized server \cite{faigl2012goal,yan2023mui}. These methods utilize global information to achieve optimal task allocation but require reliable communication and may cause delays in exploration planning. 
To improve the robustness against transmission loss, decentralized exploration task allocation strategies based on pairwise interaction are proposed in \cite{racer} and \cite{ethz}. The main drawback of pairwise interaction is that only two robots are considered, which will become inefficient when the number of robots increases.
Supported by Voronoi diagrams, the works in \cite{hu2020voronoi} and \cite{bi2023cure} divide the Euclidean space into cells for the closest robot to explore. However, the shortest exploration distance can not be guaranteed in complex scenarios since obstacles are not taken into account.

In this paper, we propose a novel decentralized multi-UAV exploration method that features efficient communication and reasonable task allocation.
To facilitate data transmission, UAVs jointly maintain a multi-robot dynamic topological graph (MR-DTG) consisting of traversable paths in the explored space and explorable regions of interest. UAVs exchange their newly constructed parts of MR-DTG to synchronize the global exploration information. Based on the fast-sharing MR-DTG, we devise a hierarchical algorithm that plans exploration actions on two levels, i.e., the local and global levels. Both levels construct graph Voronoi diagrams on MR-DTG to avoid multiple UAVs exploring a region at the same time.
Our method is validated in both the simulation and real-world environments. The simulation results show that our method outperforms the state-of-the-art method in terms of exploration time (up to 38.3\% shorter) and communication volume (up to 95.5\% lower). A sample of multi-UAV exploration is shown in Fig.~\ref{example}, wherein the environment is explored by 15 UAVs with the proposed method. 

The contributions of this work are summarized as follows:
\begin{itemize}
	
\item A newly designed multi-robot dynamic topological graph, which can significantly reduce the communication volume while maintaining the necessary information required by exploration planning.
\item A hierarchical multi-UAV exploration framework is proposed utilizing graph Voronoi partition on MR-DTG rather than Euclidean Voronoi partition of the space, which enables reasonable task allocation.
\item Extensive simulation and real-world experiments validate that our method outperforms the state-of-the-art method in terms of communication volume and exploration efficiency. We will release the source code to benefit the community\footnote{https://github.com/NKU-MobFly-Robotics/GVP-MREP}.

\end{itemize}

\section{RELATED WORK}
Autonomous exploration using one mobile robot has been widely investigated over the past decades.
Existing methods can be roughly classified as sampling-based and frontier-based methods. Sampling-based methods randomly sample feasible viewpoints in the feasible space to extend rapidly-exploring random trees and evaluate the exploration gain of each tree node for exploration goal determination \cite{rh_nbvp}. It is proven that this method is easy to get stuck in local areas. To alleviate this problem, many methods construct the history graph that connects explored spaces to guide the UAV to informative areas when the local sampling fails \cite{witting2018history, zhang2021novel, real}.
However, the maintenance of a dense history graph could be time-consuming, and the degree of the graph node is typically low, which might result in long detours because of missing shortcuts.
Frontier-based methods guide the robotic exploration using frontiers, wherein a frontier is the boundary between the known and unknown spaces \cite{yamauchi1997frontier, duberg2022ufoexplorer, yu2023echo}. In frontier-based explorations, the primary problem that should be solved is the frontier visiting sequence. Existing methods usually employ the traveling salesman problem (TSP) to obtain the best frontier visiting sequence \cite{meng2017two, cao2021tare, zhao2023autonomous}. However, the computational cost for solving TSP grows dramatically as the environment size increases. 


Different from single-robot exploration, multi-robot exploration needs to solve the problems of exploration-oriented communication and task allocation.
To synchronize the exploration information, the work in \cite{yu2021smmr} shares a point cloud map in the form of submaps. However, the point cloud accumulation leads to large communication burden and low communication frequency. Differently, Zhou et al. \cite{racer} utilized online hgrid decomposition of the exploration space to represent exploration tasks and transfered hgrid to share exploration information. Since hgrid does not contain path information between frontiers, the method needs to exchange occupancy map information in the form of chunks for path planning, which still requires a large communication burden. Instead of transferring the occupancy map, the works in \cite{MR-TopoMap} and \cite{best2022resilient} transfer topological graphs to synchronize exploration information, which can significantly reduce communication volume. However, the visibility between connected vertices is required, making the graph denser and denser in large and cluttered environments.

On the other hand, reasonable task allocation is essential for improving the multi-robot exploration efficiency. In the early work \cite{YAMAUCHI1999111}, each robot selects the frontier closest to itself as the next target. This strategy could be inefficient as multiple robots may explore the same frontier. Recently, pairwise cooperation strategies have been investigated. In the work \cite{racer}, the exploration cost of each pair of UAVs is minimized by solving a capacitated vehicle routing problem (CVRP). Bartolomei et al. \cite{ethz} enables UAVs to switch status between explorer and collector, which aims to clear small portions of unknown space. The pairwise strategy can minimize the coverage path but ignores the cooperation with other UAVs of the team.
Another type of method sends robots to the regions divided by Voronoi partition to avoid duplicated exploration \cite{hu2020voronoi, bi2023cure}. These methods can make robots distributed more evenly over the environment, but the shortest exploration distance can not be guaranteed in complex scenarios since obstacles are not considered.

Motivated by the above limitations, we propose a multi-UAV exploration method that utilizes MR-DTG to maintain and synchronize the exploration information. Compared with existing methods, our communication data type can significantly reduce the communication volume and ensure fast information sharing. Moreover, the graph Voronoi partition on MR-DTG can achieve reasonable task allocation for improving exploration efficiency in cluttered environments.

\begin{figure}[tp]
	\centerline{\includegraphics[width=0.43\textwidth]{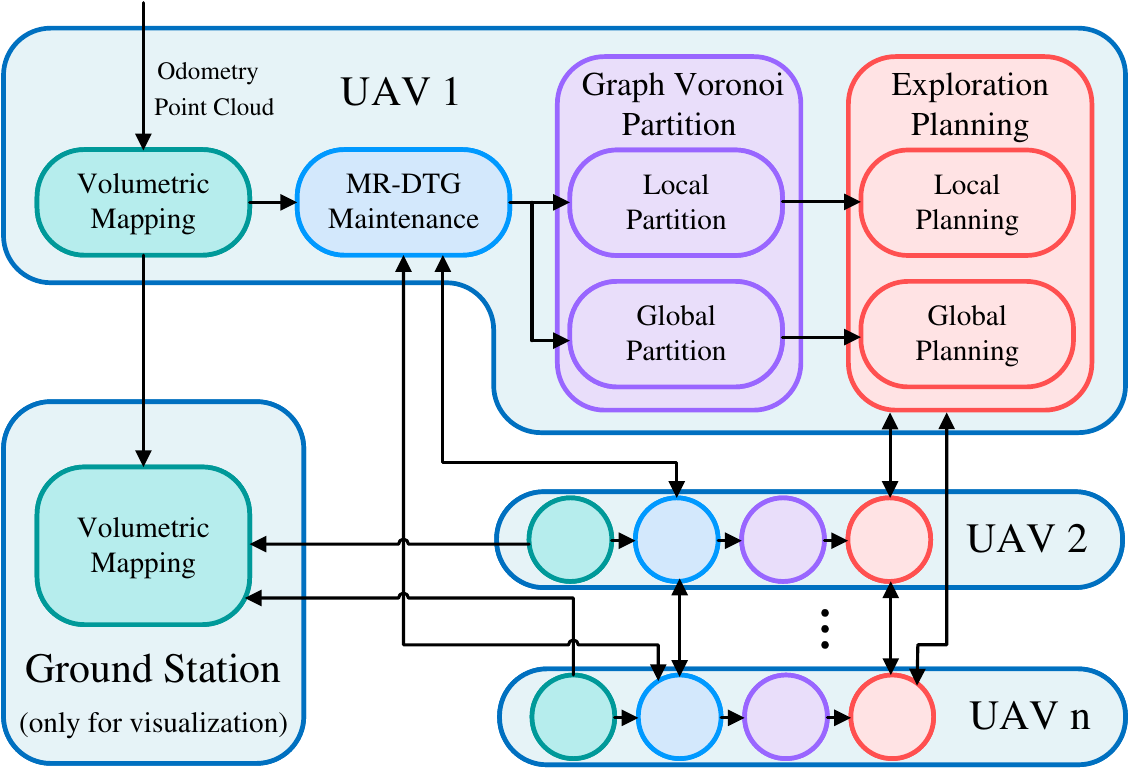}}                                     \caption{Overview of the proposed multi-UAV exploration system.}
	\label{sys}
\end{figure}

\section{METHODOLOGY}
\subsection{System Overview}
The system overview of the proposed method is illustrated in Fig.~\ref{sys}. The proposed multi-UAV system aims to build a complete volumetric map of the target space. Each UAV localizes itself using on-board sensors. The volumetric mapping module integrates odometry and point cloud to update the 3-D volumetric map, which is incrementally sent to the ground station for visualization. 
After that, the MR-DTG maintenance module updates MR-DTG according to the newly constructed volumetric map. Meanwhile, the updated part of MR-DTG is broadcast to other UAVs to synchronize the exploration information (Sect.\ref{subsec:MR-DTG}).
To facilitate exploration task allocation, the graph Voronoi partition module will partition the obtained MR-DTG at the local and global levels respectively with a constant frequency (Sect.\ref{subsec:GVP}).
If there are explorable regions allocated to the UAV, the local exploration planning module will find the exploration target for it. Otherwise, the global exploration planning module will guide the UAV to another target according to the result of the global graph Voronoi partition on MR-DTG (Sect.\ref{subsec:planning}).
The exploration will be terminated if there is no explorable region in MR-DTG.

\subsection{Fast-Sharing Multi-Robot Dynamic Topological Graph} \label{subsec:MR-DTG}
The explorable region and the traversable path information are crucial for exploration planning. To reduce the communication volume while maintaining the crucial information, we propose a novel multi-robot dynamic topological graph (MR-DTG). 
MR-DTG, $\mathbb G = (\mathcal{V}, \mathcal{E})$, is composed of nodes $\mathcal{V}$ and edges $\mathcal{E}$. The node set $\mathcal{V} = (\mathcal{V}_h, \mathcal{V}_e)$ contains the history nodes $\mathcal{V}_h$ and explorable regions of interest (EROI) $\mathcal{V}_e$. $\mathcal{V}_h$ is the set of history nodes that sampled from the executed trajectory of UAVs, and $\mathcal{V}_e$ is the set of EROIs representing explorable regions that have not been fully explored.  $\mathcal{E}$ is the set of edges connecting nodes in $\mathcal{V}$. In our setup, the edges are traversable paths between these history nodes and EROIs, and their weights are the corresponding path lengths. 
In what follows, we will introduce the maintenance of MR-DTG and the multi-UAV communication process.

\subsubsection{History Node Generation}

\begin{figure}[tp]
	\centerline{\includegraphics[width=0.43\textwidth]{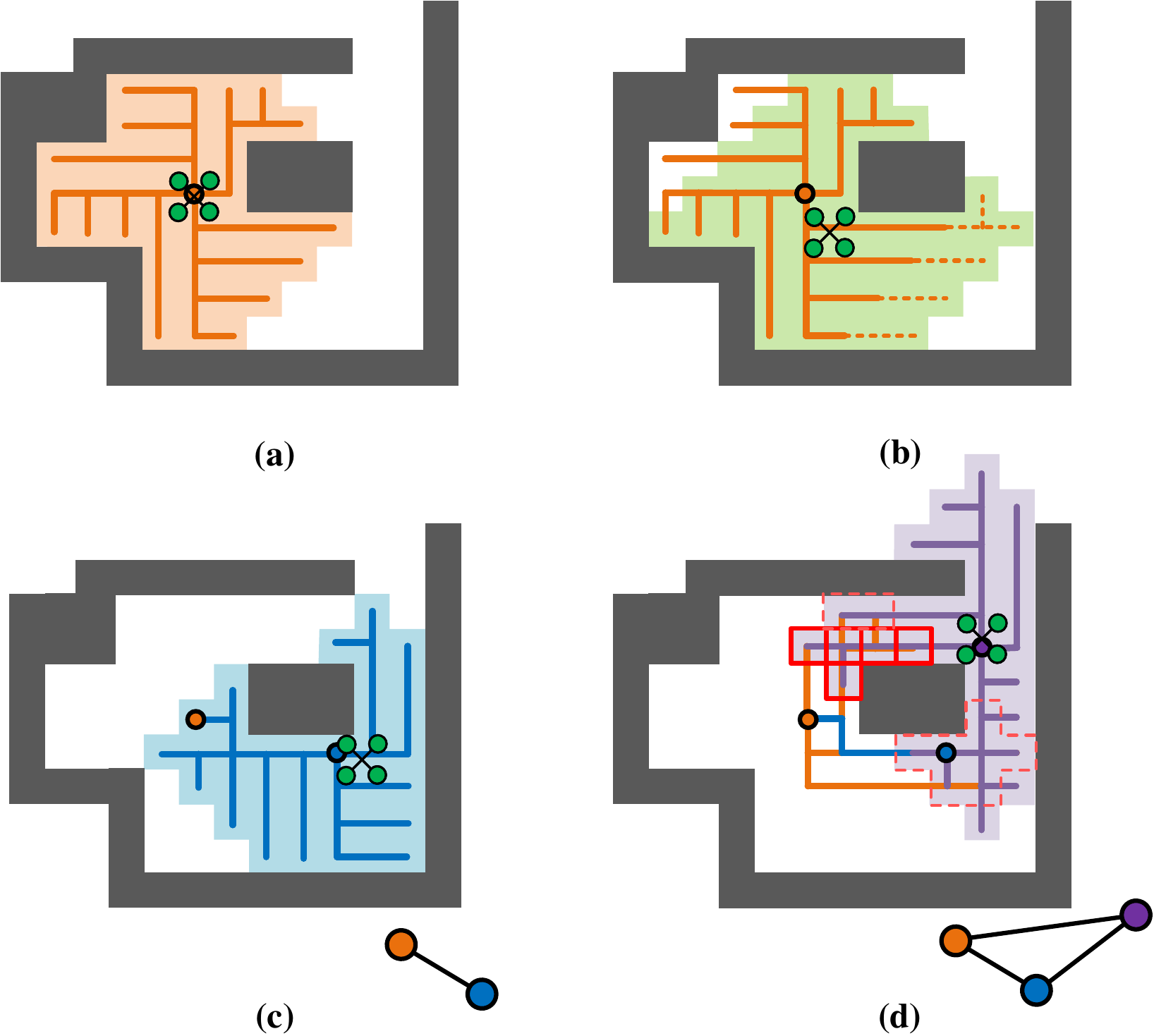}}                                     \caption{The generation of the history node in MR-DTG.
		(a) Each history node and UAV maintains a Dijkstra tree to calculate the shortest path from the surrounding voxel to it.
		(b) The current history node will expand its Dijkstra tree to cover the space that the UAV's Dijkstra search covers (green background).
		(c) A new history node will be generated when the UAV is far away from history nodes. The new history node will be connected to nodes that are covered by its Dijkstra tree.
		(d) History nodes are connected through handshakes (the orange and purple nodes ``handshake" in voxels surrounded by dotted lines). Any one of the shortest paths corresponding to the handshake voxels (voxels surrounded by solid lines) will be stored in the edge between these history nodes.
		}
	\label{hn}
\end{figure}

In this work, history nodes and the shortest paths between them constitute the sparse representation of the explored space, which are crucial for efficiently synchronizing the traversable path information between UAVs. To generate appropriate history nodes, a Dijkstra search within a maximum Manhattan distance $d_{max}$ starting from the UAV position is executed during exploration. A new history node will be created at the position of the UAV and added to $\bm{\mathcal{V}}_h$ if no history node is found in the searched space (see Fig.~\ref{hn}(a)) or the distance (calculated by the Dijkstra search) from the UAV to all its surrounding history nodes is larger than a threshold $p_{th} \leq d_{max}$ (see Fig.~\ref{hn}(c)). Whenever a new history node is generated, a \emph{Dijkstra tree} whose branches are the shortest paths to its surrounding voxels will be constructed concurrently.
If the distance from the UAV to the closest history node is less than $p_{th}$, this closest history node will expand its Dijkstra tree to cover the space that the UAV's Dijkstra search covers (see Fig.~\ref{hn}(b)).
Then, edges between history nodes will be built to enable path planning and task allocation. If a history node is directly connected by a Dijkstra tree of another history node, an edge connecting these two nodes is created by retrieving a path from this history node utilizing this Dijkstra tree (see Fig.~\ref{hn}(c)). To further increase the degree of history nodes, two history nodes that have ``handshake'' voxels, i.e., the voxels concurrently connected by the two Dijkstra trees starting from these two history nodes, are connected. The paths connecting these two history nodes can be obtained by retrieving from the ``handshake'' voxels on their Dijkstra trees. We select the shortest path as the edge between these two history nodes (see Fig.~\ref{hn}(d)).

\subsubsection{Explorable Region of Interest}
\begin{figure}[tp]
	\centerline{\includegraphics[width=0.33\textwidth]{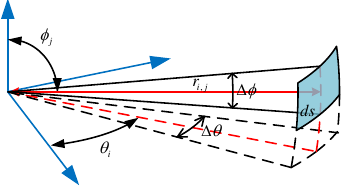}}                                     \caption{A microelement of frontier area $ds$ that the collision-free ray reaches at a distance $r_{i,j}$ in the direction $(\theta_i, \phi_j)$. The angular space within the viewpoint's FoV is evenly divided in the yaw and pitch directions with sizes $\Delta\theta$ and $\Delta\phi$ respectively.}
	\label{aep}
\end{figure}
To efficiently identify the explorable regions and further facilitate multi-UAV communication, we devise a multi-robot-exploration-oriented data structure named the explorable region of interest (EROI). Firstly, the whole exploration space is evenly divided into cubes (i.e., EROIs) with unique indexes (IDs) before the exploration. Meanwhile, candidate viewpoints for an EROI are generated by uniform sampling in the cylindrical space whose origin is the corresponding EROI center. The state of an EROI belongs to one of the following states: $inactiveR$, $activeR$, and $deadR$. Similarly, the state of a viewpoint belongs to one of the following states: $inactiveV$, $activeV$, and $deadV$. The yaw angle of each viewpoint is towards the EROI center. 
Initially, the states of all the EROIs and their viewpoints are $incativeR$ and $inactiveV$, respectively. Once any voxels in an EROI are observed, the state of the EROI will be switched to $activeR$. Then, for all $activeR$ EROIs, the information gains of its collision-free $activeV$ and $inactiveV$ viewpoints will be calculated. Specifically, inspired by \cite{AEP} for each viewpoint, the information gain is calculated by casting lines outward from the viewpoint and adding up all the areas of frontiers hit by the lines. Note that our method does not maintain frontiers explicitly. Instead, we regard unobserved voxels hitting by the ray as frontiers. A microelement of the frontier area $ds$ that the ray reaches at a distance $r_{i,j}$ in the direction $(\theta_i, \phi_j)$ is depicted in Fig.~\ref{aep}. The frontier area hit by this ray with a discrete angular region of $(\Delta \theta, \Delta \phi )$ is computed by
\begin{equation} 
	\begin{aligned}
		ds(r_{i,j},\theta_i,\phi_j) & = \int_{\theta_i -\frac{\Delta\theta}{2}}^{\theta_i + \frac{\Delta\theta}{2}}\int_{\phi_j - \frac{\Delta\phi}{2}}^{\phi_j + \frac{\Delta\phi}{2}}r_{i,j}^2\sin(\phi) d\phi d\theta \\ & = 2r_{i,j}^2\Delta\theta\sin(\phi_i)\sin(\Delta\phi/2).
	\end{aligned}
\end{equation}
If there is no frontier in the maximum detection range $r_{max}$, $r_{i,j}$ will be set to 0.
The viewpoint with a yaw angle $\theta_v$, its information gain is calculated as the sum of the information gains in each discrete direction in the field-of-view (FoV):
\begin{equation} 
	GViewpoint(\theta_v)=\sum\nolimits_{j=0}^{m}\sum\nolimits_{k=0}^{n}ds(\theta_j,\phi_k)
\end{equation}
where
\begin{equation} 
	m = \frac{FoV_{\theta_{left}}-FoV_{\theta_{right}}}{\Delta\theta} ,\  n = \frac{FoV_{\phi_{up}}-FoV_{\phi_{down}}}{\Delta\phi} 
\end{equation}
\begin{equation} 
	\theta_j=\theta_v+FoV_{\theta_{right}}+j\Delta\theta ,\ \phi_k=FoV_{\phi_{down}}+k\Delta\phi
\end{equation}
If the information gain $g(\theta_v)$ exceeds the threshold $g_{th}$, the state of the viewpoint will be switched to $activeV$. Otherwise, the state of the viewpoint is switched to $deadV$. 
To add the $activeR$ EROI to MR-DTG, all $activeV$ viewpoints of the $activeR$ EROI will attempt to connect the history nodes of MR-DTG, and the shortest path will be selected as the edge connecting $activeR$ EROI and MR-DTG. 
Thanks to the reserved Dijkstra trees, the paths between $activeV$ viewpoints and history nodes can be queried directly. To ensure the connecting edge between one $activeR$ EROI and MR-DTG is always the shortest, the above process will be repeated continuously during the exploration. In addition, if the state of the viewpoint previously connected to the history node is switched to $deadV$, the corresponding edge will be deleted. Note that each EROI can only connect to at most one history node. When the coverage rate of an EROI exceeds the designated threshold $e_{th}$ or the states of all its viewpoints are $deadV$, the EROI is considered to be fully explored, and its state will switch to $deadR$. All $deadR$ EROIs will be deleted from the MR-DTG. In addition, during the exploration, we continuously execute the Dijkstra search started from the UAV's position and directly connect all $activeR$ EROIs and history nodes within $d_{max}$ to the UAV to improve the local exploration efficiency. It is worth noting that the EROI benefits the multi-robot exploration in two aspects. First, our method is free from the time-consuming frontier detection and clustering. Second, only the IDs and states of the EROIs need to be broadcast to synchronize the exploration information. 

\begin{figure}[tp]
	\centerline{\includegraphics[width=0.46\textwidth]{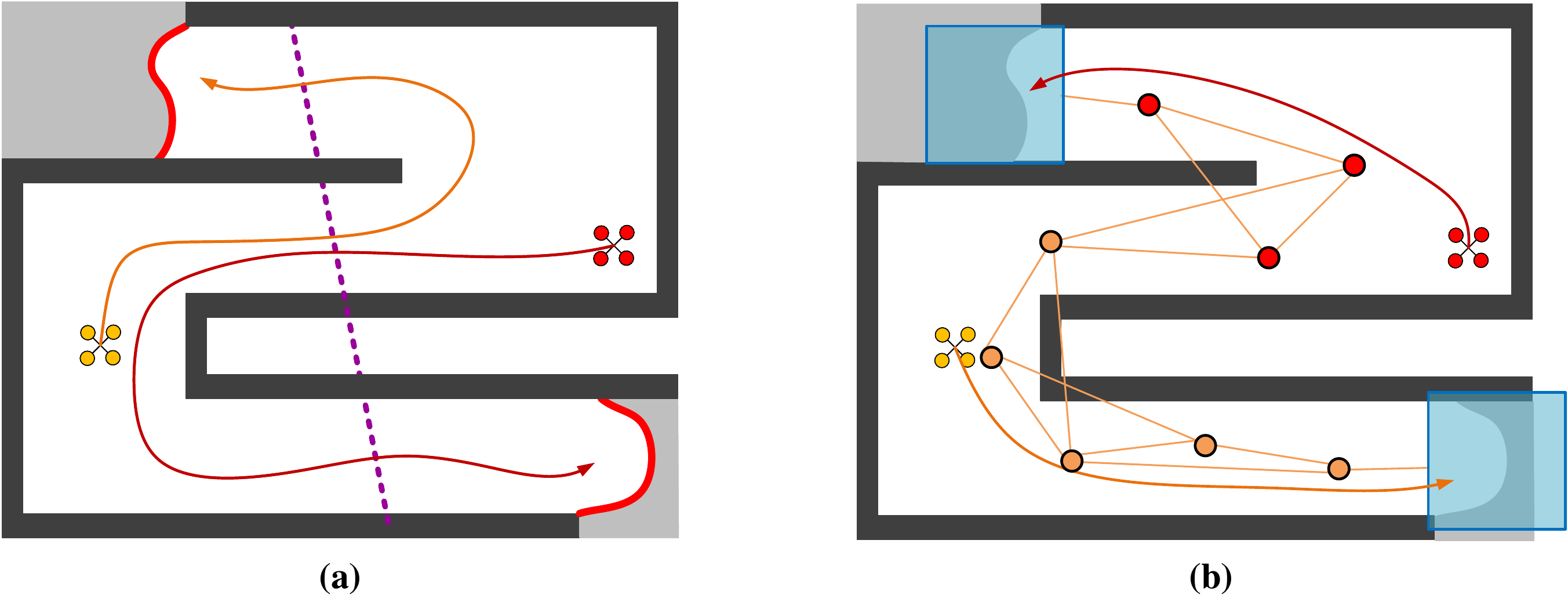}}   
	\caption{Comparison of Voronoi partition of Euclidean space and graph Voronoi partition on MR-DTG.
		(a) The Voronoi partition of Euclidean space assigns the upper target to the yellow UAV and the lower target to the red UAV. Due to the existence of obstacles, both UAVs must take a long detour to reach the target.
		(b) Graph Voronoi partition on MR-DTG takes into account the traversable path length between nodes, assigning EROI to UAVs with the lowest exploration motion cost.}
	\label{VP-comparison}
\end{figure}

\subsubsection{Multi-Robot Communication} 
In order to synchronize the exploration information quickly, UAVs only need to exchange the MR-DTG information incrementally, which can significantly reduce the communication volume. Specifically, when the graph structure of MR-DTG changes, newly generated history nodes, the states and IDs of the updated EROIs and their corresponding viewpoints, and the updated edges will be transferred between UAVs. In addition, the nodes of MR-DTG that directly connect with the UAV and their distances to the UAV will also be broadcast for the following task allocation.

\subsection{Voronoi Partition on MR-DTG}\label{subsec:GVP}
Classic Euclidean Voronoi partition for multi-robot exploration \cite{hu2020voronoi, bi2023cure} typically divides space into disjoint regions whose Euclidean distance is closest to each robot and guides robots to explore different regions separately without interfering with each other. However, classic Euclidean Voronoi partition may cause unreasonable task allocation because obstacles are not considered (see Fig.~\ref{VP-comparison}(a)). To avoid this problem, we propose constructing local and global graph Voronoi Diagrams \cite{gvp} on MR-DTG to allocate the truly closest exploration targets to each UAV. Both the local and global partition results are utilized to guide the subsequent exploration planning. Thanks to real-time communication, each UAV will always have the same MR-DTG and the connection relationship between MR-DTG and other UAVs so that UAVs can allocate exploration tasks quickly and decentralizedly without cooperating with other UAVs while getting the same allocation results. 

\begin{figure}[tp]
	\centerline{\includegraphics[width=0.43\textwidth]{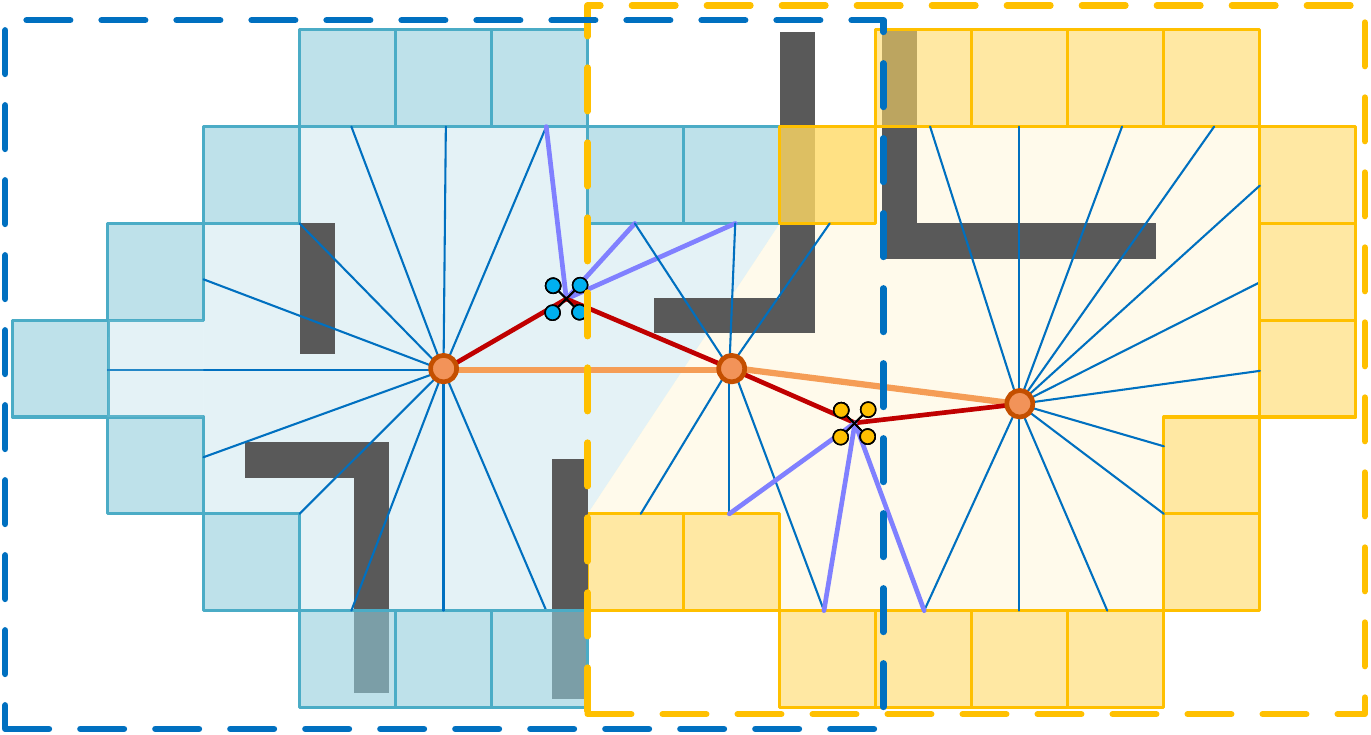}}   
	\caption{Local Voronoi partition on MR-DTG. Each drone constructs its local search graph before partition. For the blue UAV, the local MR-DTG corresponds to the elements in the blue dotted line, including the left and middle history nodes, two UAV nodes, and their connected EROIs. After partition, the blue EROI is assigned to the blue UAV for exploration.
	 }
	\label{local_voronoi_partition}
\end{figure}

\subsubsection{Local Voronoi Partition on MR-DTG}
The local Voronoi partition is utilized to prevent UAVs from exploring the same EROI concurrently. Fig.~\ref{local_voronoi_partition} shows an example of local Voronoi partition on MR-DTG. For brevity, we take the local graph Voronoi partition on MR-DTG conducted by the $i$-th UAV as an example. First, a local graph $\mathcal{G}_l^i$ for the $i$-th UAV performing local Voronoi partition will be constructed. Its nodes $\mathcal{N}_l^i$ are composed of history nodes $\mathcal{N}_{h}^i$, EROIs $\mathcal{N}_{e}^i$, and UAVs $\mathcal{N}_{u}^i$. Wherein, $\mathcal{N}_{h}^i$ consists of history nodes directly connected to the $i$-th UAV. $\mathcal{N}_{e}^i$ consists of EROIs that directly connected by the $i$-th UAV and connected by history nodes in $\mathcal{N}_{h}^i$. $\mathcal{N}_{u}^i$ consists of the UAVs directly connected with the EROIs in $\mathcal{N}_{e}^i$ and connected with history nodes in $\mathcal{N}_{h}^i$. Then, the parallel Dijkstra algorithm \cite{gvp} is conducted starting from $\mathcal{N}_{u}^i$ for the graph Voronoi partition. According to the results of the parallel Dijkstra search, the local EROIs $\mathcal V_{le}^i\subseteq \mathcal{N}_{e}^i$ that with the shortest paths to the $i$-th UAV are allocated to the $i$-th UAV for local exploration.

\subsubsection{Global Voronoi Partition on MR-DTG}
Global graph Voronoi partition is used to allocate all history nodes of MR-DTG to the UAVs, and then each UAV will obtain an exploration region. Specifically, we first construct a global graph $\mathcal{G}_g$ whose nodes $\mathcal{N}_g$ are composed of all the history nodes $\mathcal{N}_{h}$ of MR-DTG and all the positions of UAVs $\mathcal{N}_u$. Similar to the local Voronoi partition, we take the global graph Voronoi partition on MR-DTG conducted by the $i$-th UAV as an example. Supported by the parallel Dijkstra search, which starts from the $i$-th UAV, the global history nodes that with the shortest paths to the $i$-th UAV are allocated to the $i$-th UAV. Therefore, it will get a set of closest history nodes $\mathcal{V}_{gh}^i \subseteq \mathcal{N}_{h}$ for global exploration. Due to the fact that the exploration information has always been synchronized, that is, each UAV will always have the same MR-DTG. Therefore, the global graph Voronoi partition is also only required to be conducted by each UAV itself.
 

\subsection{Exploration Planning}  \label{subsec:planning}
After the exploration task allocation using the graph Voronoi partition on MR-DTG, a hierarchical planning framework is used to find the exploration targets and generate exploration paths for the UAVs. The exploration planning process for the $i$-th UAV is illustrated in Alg. \ref{alg:planning}.
\begin{algorithm}[tp]
	\caption{Hierarchical planning for the $i$-th UAV}\label{alg:planning}
	\begin{algorithmic}[1]
		\renewcommand{\algorithmicrequire}{ \textbf{Input:}}
		\REQUIRE
		$\mathcal{V}_{le}^i, \mathcal{V}_{gh}^i, \mathbb G$
		\renewcommand{\algorithmicrequire}{ \textbf{Output:}}
		\REQUIRE
		target EROI $e_t^i$
		\IF{$\mathcal{V}_{le}^i \ne \emptyset $} 
		\STATE $e_t^i \gets$ GetClosestEROI$(\mathcal{V}_{le}^i, \mathbb{G})$
		\ELSE
		\STATE $\mathcal{V}^{i}_{ha}$ $\gets$ GetHnWithEROI($\mathcal{V}_{gh}^i$)
		\STATE$Gain_h$ $\gets$ GetGHs($\mathbb{G}$)
		\IF{$\mathcal{V}^{i}_{ha} \ne \emptyset $}
		\STATE $v_{ht}^i \gets$ GetClosestHn$(\mathcal{V}^{i}_{ha},\mathbb{G})$
		\STATE  $e_t^i \gets$ GetClosestEROI$(v_{ht}^i, \mathbb{G})$
		\ELSIF{$\exists g \in Gain_h, g > 0$}
		\STATE $v_{ht}^i \gets$  GetHighestGainHn$(\mathcal{V}^{i}_{ha},Gain_h)$
		\STATE $e_t^i \gets$ GetClosestEROI$(v_{ht}^i,\mathbb{G})$
		\ELSE
		\STATE exploration finishes
		\ENDIF
		\ENDIF
	\end{algorithmic}
\end{algorithm} 

\subsubsection{Local Planning}  \label{subsubsec:local_planning}
In local exploration planning (Alg. \ref{alg:planning}, lines 1-2), a greedy strategy is adopted, that is, an EROI will be explored by the UAV closest to it. Thanks to the Dijkstra search starting from the position of the $i$-th UAV, which has been continuously performed during the exploration for MR-DTG maintenance, the path lengths between the $i$-th UAV and its directly connected EROIs are obtained in advance. For EROI $\in \mathcal V_{le}^i$ that does not directly connect to the UAV, the path will be found by the Dijkstra search on MR-DTG. 

\subsubsection{Global Planning}
When the $i$-th UAV has no local EROI to explore, it will start global exploration planning to find a new exploration target (Alg. \ref{alg:planning}, lines 3-15). If there are EROIs connected with $\mathcal V_{gh}^i$, the $i$-th UAV will select the nearest EROI connected with a history node of $\mathcal V_{gh}^i$ as the next exploration target (Alg. \ref{alg:planning}, lines 6-8). Otherwise, it means that the UAV completes its own exploration task and starts to look for new exploration targets in the exploration regions of other UAVs to share the exploration burden. Specifically, the $i$-th UAV will first calculate the information gain of all history nodes of MR-DTG. The information gain of a history node $v_h$ is calculated by

\begin{small}
\begin{equation} \label{f:nh_gain}
GH(n_e,n_u) 
= \frac{max(n_e - \sum_{j=1}^{n_u} max(t_i-t_j,0) \cdot n_\tau ,0) + n_e \cdot g_l}{n_u + 1}e^{-\lambda t_i} 
\end{equation}
\end{small}
where $n_e$ is the number of EROIs connected to $v_h$, $n_u$ is the number of UAVs currently exploring the EROI connected with $v_h$, $t_j$ is the time cost for the $j$-th UAV reaching its exploration target, $t_i$ is the time cost for the $i$-th UAV reaching its exploration target, $n_\tau$ is an estimated parameter that describes the number of EROIs a UAV can explore per unit time, $g_l$ is a very low-value parameter to guide the UAV to $v_h$ to deal with the situation where all the EROIs are allocate to other UAVs but newly generated EROI may be assigned to it, and factor $\lambda$ is used to penalize high path costs. Here, $t_i$ is heuristically evaluated by 
\begin{equation} \label{cost}
t_i 
= \frac{Dist(UAV_i, v_h) + DistEROI(v_h)}{vel_{max}}
\end{equation}
where $Dist(UAV_i, v_h)$ is the length of the shortest path on MR-DTG, $ DistEROI(v_h)$ is the shortest edge length between $v_h$ and its EROIs, and $vel_{max}$ is the maximum velocity of the UAV. 
Then, the history node $v_{h}^i$ whose information gain is the highest is selected as the target history node $v_{ht}^i$ for the $i$-th UAV. The EROI which has the shortest distance connected to $v_{ht}^i$ is selected as the next exploration target $e_t^i$ (Alg. \ref{alg:planning}, lines 9-11). If the gains of all the history nodes are 0, the exploration finishes.

\subsubsection{Trajectory Generation}
After obtaining $e_t^i$ by the hierarchical exploration planning, the $i$-th UAV will move towards the viewpoint of $e_t^i$. If the viewpoint of $e_t^i$ is in the free space explored by the $i$-th UAV, a path to it will be searched to shorten the moving distance. Otherwise, the $i$-th UAV will move following the path to the viewpoint provided by MR-DTG to avoid collision. To further speed up the exploration, MINCO trajectory \cite{minco} is utilized to obtain an efficient and smooth collision-free trajectory for UAV tracking. 

After finding the local or global exploration target, the ID of $e_t^i$ and history node connected to $e_t^i$, as well as $t_i$ are broadcast and used for other UAVs' exploration planning.

\section{EXPERIMENTS}

\subsection{Inplementation Details}
In order to evaluate our method thoroughly, simulation and real-world experiments are conducted. For simulation trials, all involved algorithms are implemented in C++ on a computer that runs Robot Operating System (ROS) with an AMD Ryzen 7 3700X CPU. Both the proposed method and the competitive method are performed in a realistic simulator, Gazebo. The multi-UAV system used in the simulator is provided by Rotors \cite{furrer2016rotors}. Each UAV is equipped with a RGB-D camera whose FoV parameters are set as follows: ${FoV}_{\theta_{left}}=57.3\,^\circ$, ${FoV}_{\theta_{right}}=-57.3\,^\circ$, ${FoV}_{\phi_{up}}=45.0\,^\circ$, ${FoV}_{\phi_{down}}=-45.0\,^\circ$ and $r_{max}=5.0\,\rm m$. We set $d_{max}=10.0\,\rm m$ and $p_{th}=5.5\,\rm m$ for MR-DTG construction. To evaluate viewpoints and EROIs, we set $(\Delta \theta, \Delta \phi) = (7.5\,^\circ, 7.5\,^\circ)$, $g_{th} = 1.3\,\rm m^2$ and $e_{th} = 85\,\%$. For task allocation, we have $n_\tau = 0.2$, $g_l = 0.08$, $\lambda = 0.1$, and $vel_{max} = 1.5\,\rm{m/s}$. In the real-world experiment, the UAV platform employed in the multi-UAV system is equipped with a Livox MID360 LiDAR (see Fig. 8(b)). The UAV testbed is tailored for outdoor exploration. The FoV parameters of the LiDAR are set as ${FoV}_{\theta_{left}}=360.0\,^\circ$, ${FoV}_{\theta_{right}}=0.0\,^\circ$, ${FoV}_{\phi_{up}}=70.0\,^\circ$, ${FoV}_{\phi_{down}}=-10.0\,^\circ$ and $r_{max}=6.0\,\rm m$. The computer mounted on each UAV is the Intel NUC with an Intel Core i7-1260p CPU. In addition, Point-LIO \cite{He_slam} is used to localize the UAV. The other parameters used in the experiment are the same as the simulation.


\subsection{Simulation Experiments}
\begin{table*}[htp]
	\centering  
	\caption{Exploration efficiency, communication volume and commputational cost.}  
	\label{exp_data} 
	\resizebox{1.0\linewidth}{!}{
	\begin{tabular}{cccccccc}  
	\toprule
		\multirow{2}{*}{\textbf{Scene}}& \multirow{2}{*}{\textbf{UAV Num}}&\multirow{2}{*}{\textbf{Method}}&\multirow{2}{*}{\textbf{Exploration Time} (s)}&\multicolumn{2}{c}{\textbf{Data Transmission} (MB)}&\multicolumn{2}{c}{\textbf{Computational Cost} (ms)}\\ 
				&& & &\textbf{Cooperation}&\textbf{Mapping}&\textbf{Partition}&\textbf{Planning}\\\midrule
				\multirow{2}{*}{Small Maze}&\multirow{2}{*}{3}&RACER \cite{racer}&115.6$\pm$18.0&\textbf{2.15$\pm$0.39}&59.42$\pm$2.30&60.12$\pm$71.99&72.30$\pm$72.77\\
				&&Proposed&\textbf{94.5$\pm$6.1}&2.32$\pm$0.20&\textbf{1.69$\pm$0.06}&\textbf{0.17$\pm$0.76}&\textbf{3.39$\pm$2.37}			\\\midrule
				\multirow{2}{*}{Small Maze}&\multirow{2}{*}{5}&RACER \cite{racer}&100.8$\pm$14.6&\textbf{5.00$\pm$0.77}&184.73$\pm$13.63&59.40$\pm$104.17&107.53$\pm$136.04\\
				&&Proposed&\textbf{69.8$\pm$7.7}&5.79$\pm$0.34&\textbf{2.56$\pm$0.12}&\textbf{0.20$\pm$0.53}&\textbf{3.61$\pm$2.74}								\\\midrule
				\multirow{2}{*}{Small Maze}&\multirow{2}{*}{10}&RACER \cite{racer}&82.1$\pm$24.8&\textbf{11.91$\pm$1.75}&537.66$\pm$28.53&51.89$\pm$98.99&44.58$\pm$81.06\\
				&&Proposed&\textbf{57.4$\pm$2.3}&21.72$\pm$1.26&\textbf{5.91$\pm$0.15}&\textbf{0.23$\pm$0.53}&\textbf{4.14$\pm$3.51}							\\\midrule
				\multirow{2}{*}{Large Maze}&\multirow{2}{*}{3}&RACER \cite{racer}&259.9$\pm$30.0&\textbf{3.64$\pm$0.64}&132.84$\pm$17.48&80.59$\pm$119.25&90.11$\pm$85.39\\									
				&&Proposed&\textbf{202.6$\pm$14.5}&4.44$\pm$0.31&\textbf{3.78$\pm$0.13}&\textbf{0.23$\pm$0.48}&\textbf{3.54$\pm$2.54}					\\\midrule
				\multirow{2}{*}{Large Maze}&\multirow{2}{*}{5}&RACER \cite{racer}&204.2$\pm$43.3&\textbf{6.13$\pm$1.36}&357.85$\pm$60.30&175.15$\pm$109.62&93.22$\pm$152.53\\
				&&Proposed&\textbf{148.2$\pm$17.7}&10.86$\pm$0.85&\textbf{5.65$\pm$0.19}&\textbf{0.23$\pm$0.16}&\textbf{3.37$\pm$2.82}							\\\midrule
				\multirow{2}{*}{Large Maze}&\multirow{2}{*}{10}&RACER \cite{racer}&191.8$\pm$34.7&\textbf{16.14$\pm$2.67}&930.42$\pm$98.01&140.03$\pm$237.46&66.73$\pm$139.14\\
				&&Proposed&\textbf{118.4$\pm$12.1}&39.32$\pm$1.35&\textbf{11.94$\pm$0.42}&\textbf{0.34$\pm$0.52}&\textbf{4.13$\pm$3.32}							\\\bottomrule
	\end{tabular}}
\end{table*}
\begin{figure}[tp]
\centering
\subfloat[]{
\centering
\begin{minipage}[htp]{0.30\linewidth}
\centering
\includegraphics[width=0.80\textwidth]{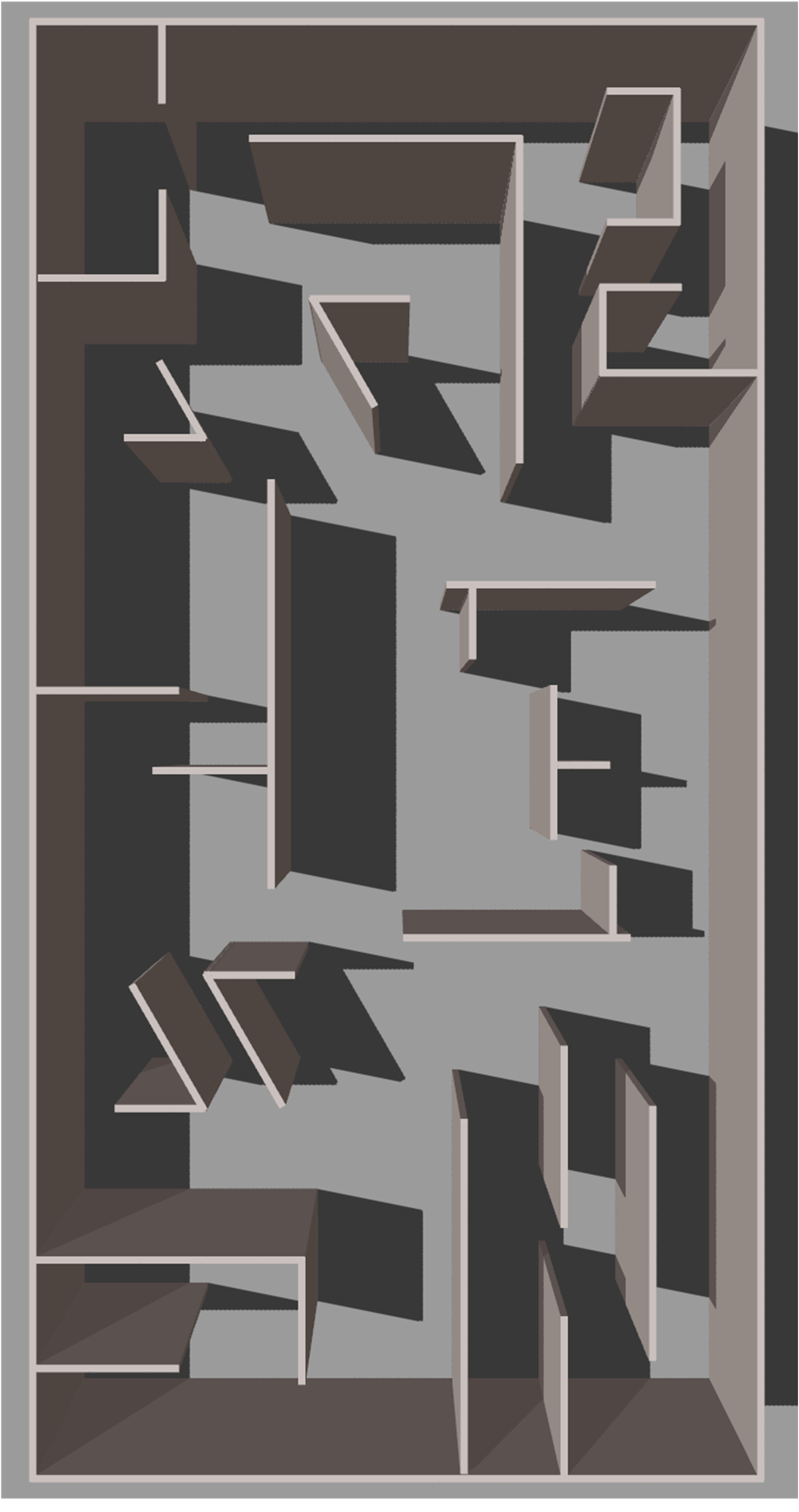}
\end{minipage}%
}%
\subfloat[]{
\centering
\begin{minipage}[htp]{0.45\linewidth}
\centering
\includegraphics[width=0.99\textwidth]{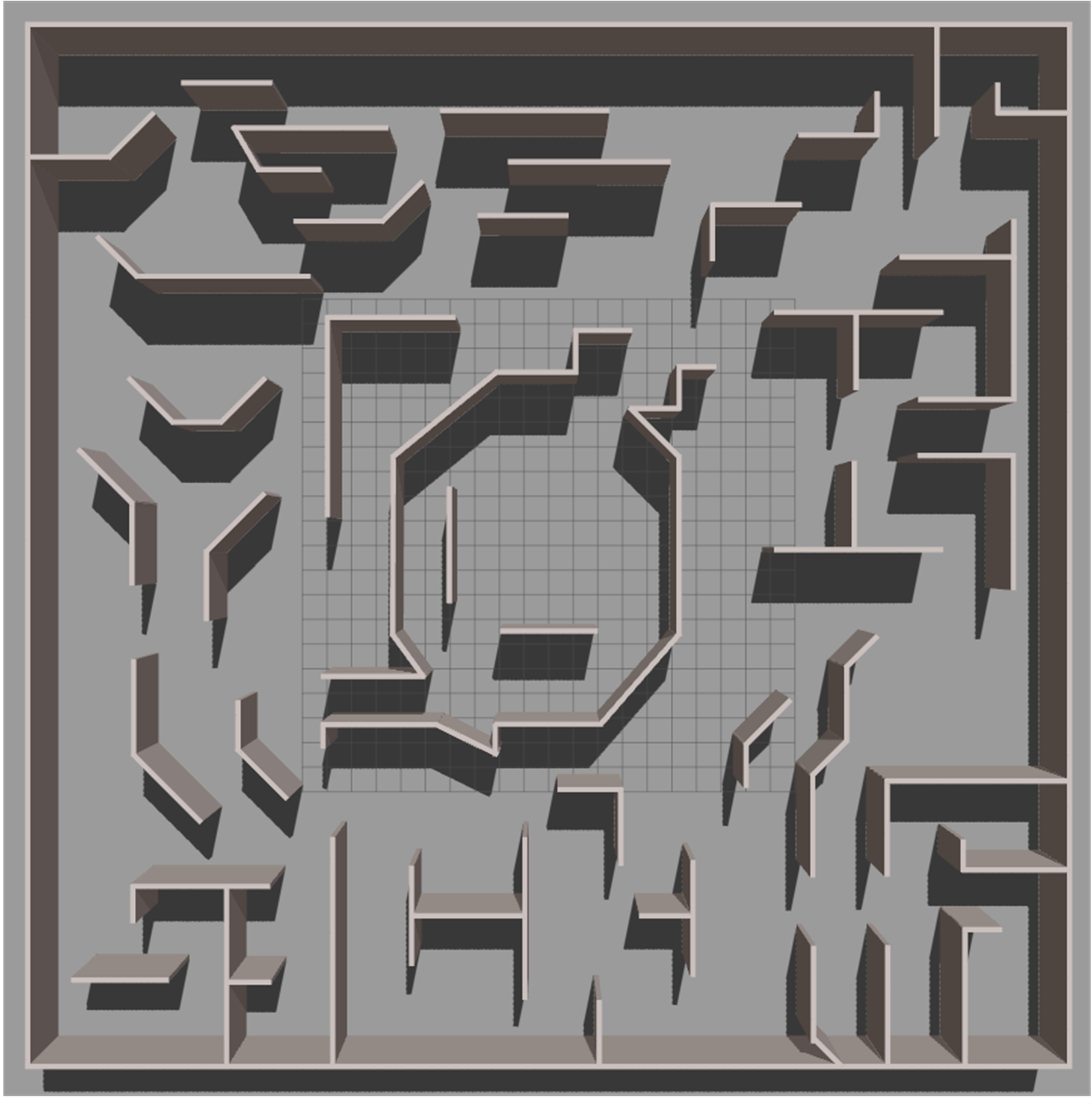}
\end{minipage}%
}

\centering
\subfloat[]{
\begin{minipage}[htp]{0.85\linewidth}
\centering
\includegraphics[width=0.90\textwidth]{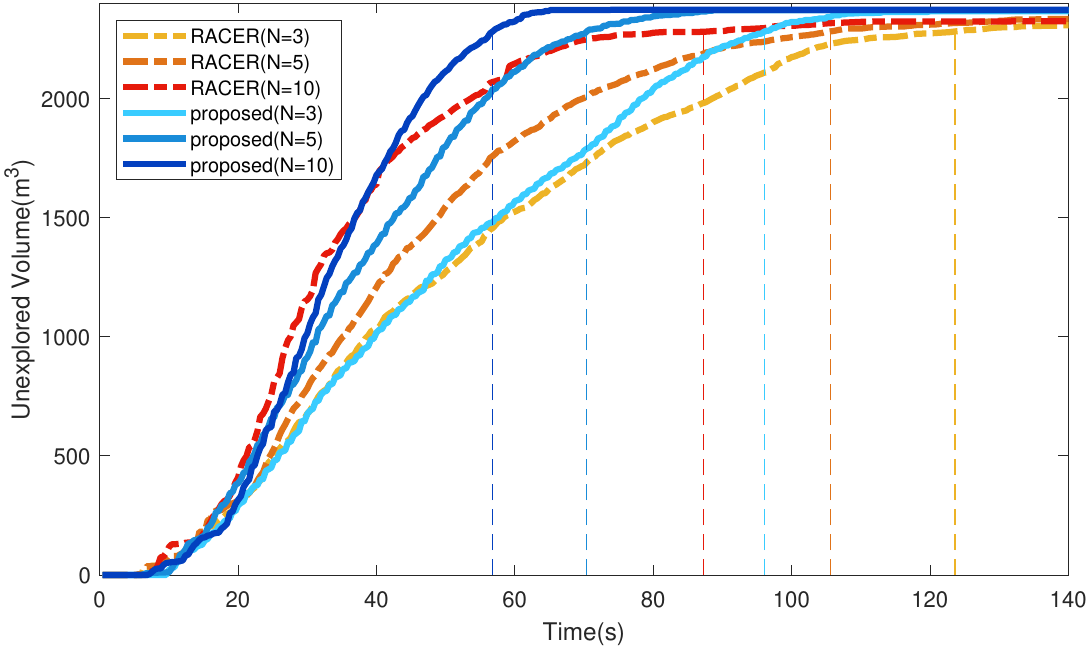}
\end{minipage}%
}

\subfloat[]{
\centering
\begin{minipage}[htp]{0.85\linewidth}
\centering
\includegraphics[width=0.90\textwidth]{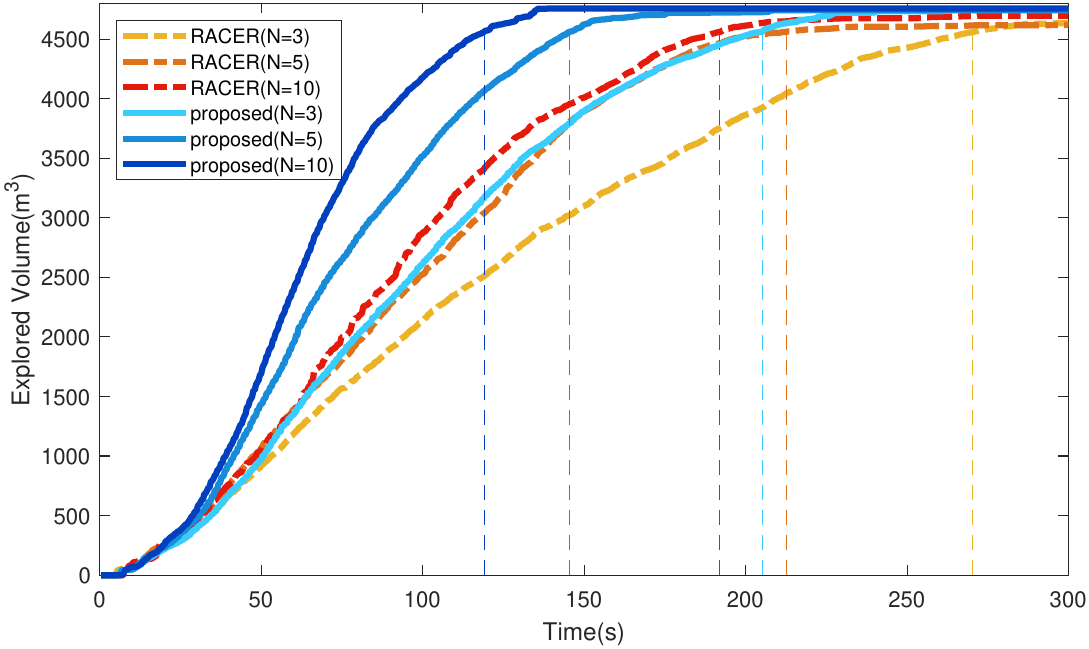}
\end{minipage}%
}%
\caption{(a) Small Maze scenario. (b) Large Maze scenario. (c) The exploration progress in Small Maze. (d) The exploration progress in Large Maze. In (c) and (d), N is the number of UAVs, the curves are the average exploration volumes and the vertical lines show the time of the average exploration volumes reaching 95\% coverage.}
\label{environments}
\end{figure}

In this section, the proposed method is evaluated in two challenging scenarios. As shown in  Fig. 7, the simulated environments are named as \emph{Small Maze} and \emph{Large Maze}. To demonstrate the performance of the proposed method, we compare it with the state-of-the-art multi-UAV autonomous exploration method, RACER \cite{racer}. Both methods are run 10 times in each scenario for statistic evaluation.  Since the involved methods have different termination criteria, we stop the exploration when the UAVs achieve 95\% coverage of the environment or a time limit is reached.

$\textbf{Small Maze}$: As illustrated in Fig.~\ref{environments}(a), the size of Small Maze is $40\times20\times3\,\rm{m^3}$. In this trial, we set the maximum run time of each method to $140$s. As shown in Table \ref{exp_data}, the proposed method saves $93.5\,\%-95.6\,\%$ communication volume than RACER by only transferring the lightweight MR-DTG instead of transferring the large-volume occupancy submaps between UAVs. In addition, our method achieves at least one order of magnitude shorter than RACER for the computational time. It is worth noting that the computational cost of RACER will decrease as the number of UAVs increases due to the reduction of the scale of CVRP. However, the pairwise cooperation strategy does not take other UAVs of the swarm into consideration, which causes lower cooperation efficiency than our graph-Voronoi-partition-based task allocation strategy. It shows that the exploration efficiency of the our method grows faster with the number of UAVs increasing than that of RACER (see Fig.~\ref{environments}(c)). Overall, as shown in Table \ref{exp_data}, the exploration time of the proposed method is about $18.3\,\%-30.8\,\%$ shorter than that of RACER with different numbers of UAVs.  


$\textbf{Large Maze}$: As illustrated in Fig~\ref{environments}(b), the size of Large Maze is $40\times40\times3\,\rm{m^3}$. In this large scenario, we set the maximum run time of each method to $300$s. It can be seen in Table \ref{exp_data}, that the computational cost of RACER increases due to the large scale of this environment. In contrast, the computational performance of our method is consistently similar to the first trial. This is because MR-DTG is sparse enough, therefore, the computational cost for graph Voronoi partition on MR-DTG is almost negligible. For the exploration efficiency, the proposed method is $22.0\,\%-38.3\,\%$ faster than RACER. Moreover, our method saves0 $94.0\,\%-95.5\,\%$ communication volume compared with RACER, which is also consistently similar to the first trial.

\subsection{Real-World Experiments}
\begin{figure}[htp]
\centering
\subfloat[]{
\centering
\begin{minipage}[htp]{0.42\linewidth}
\centering
\includegraphics[width=1.01\textwidth]{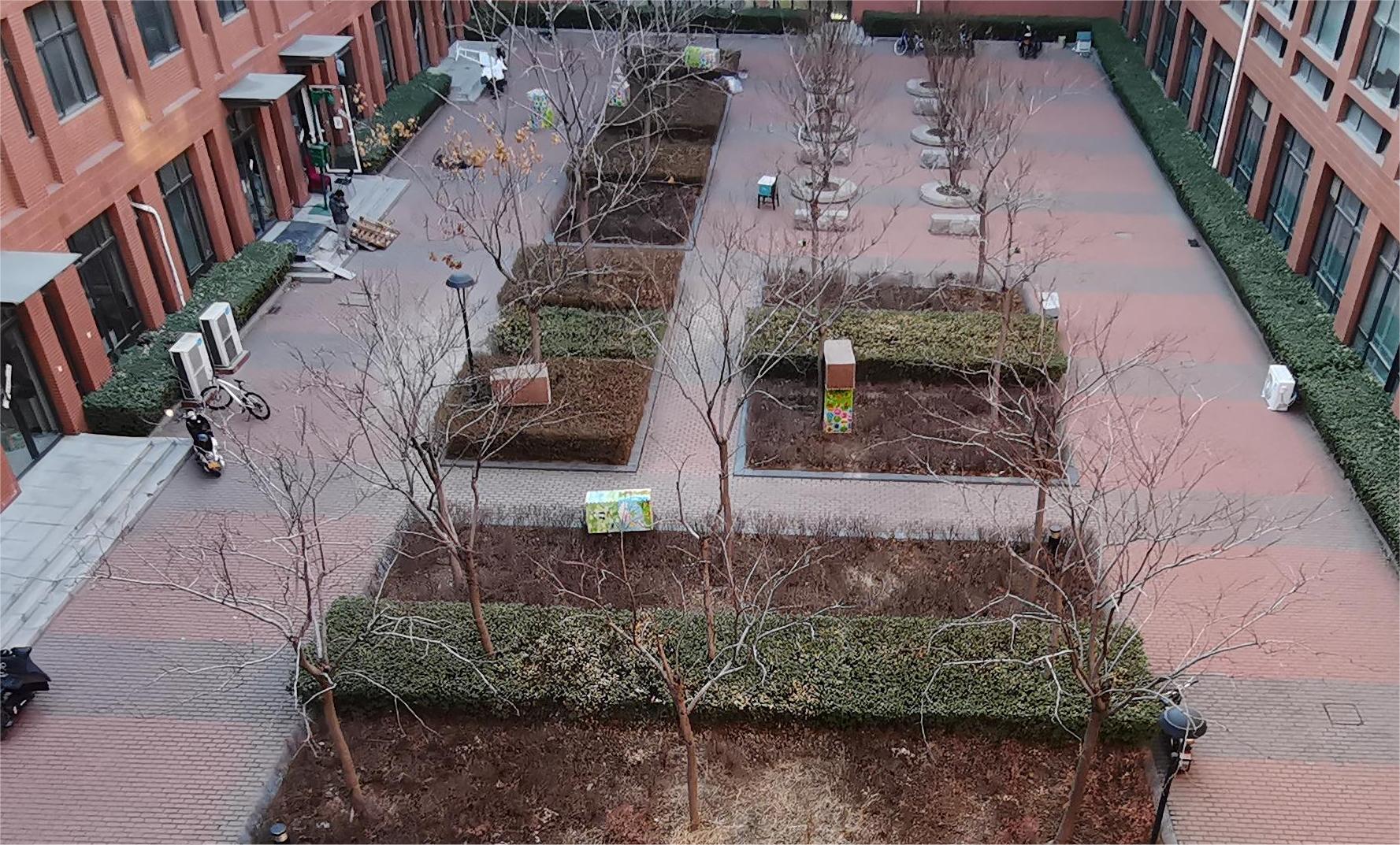}
\end{minipage}%
}%
\subfloat[]{
\centering
\begin{minipage}[htp]{0.42\linewidth}
\centering
\includegraphics[width=0.94\textwidth]{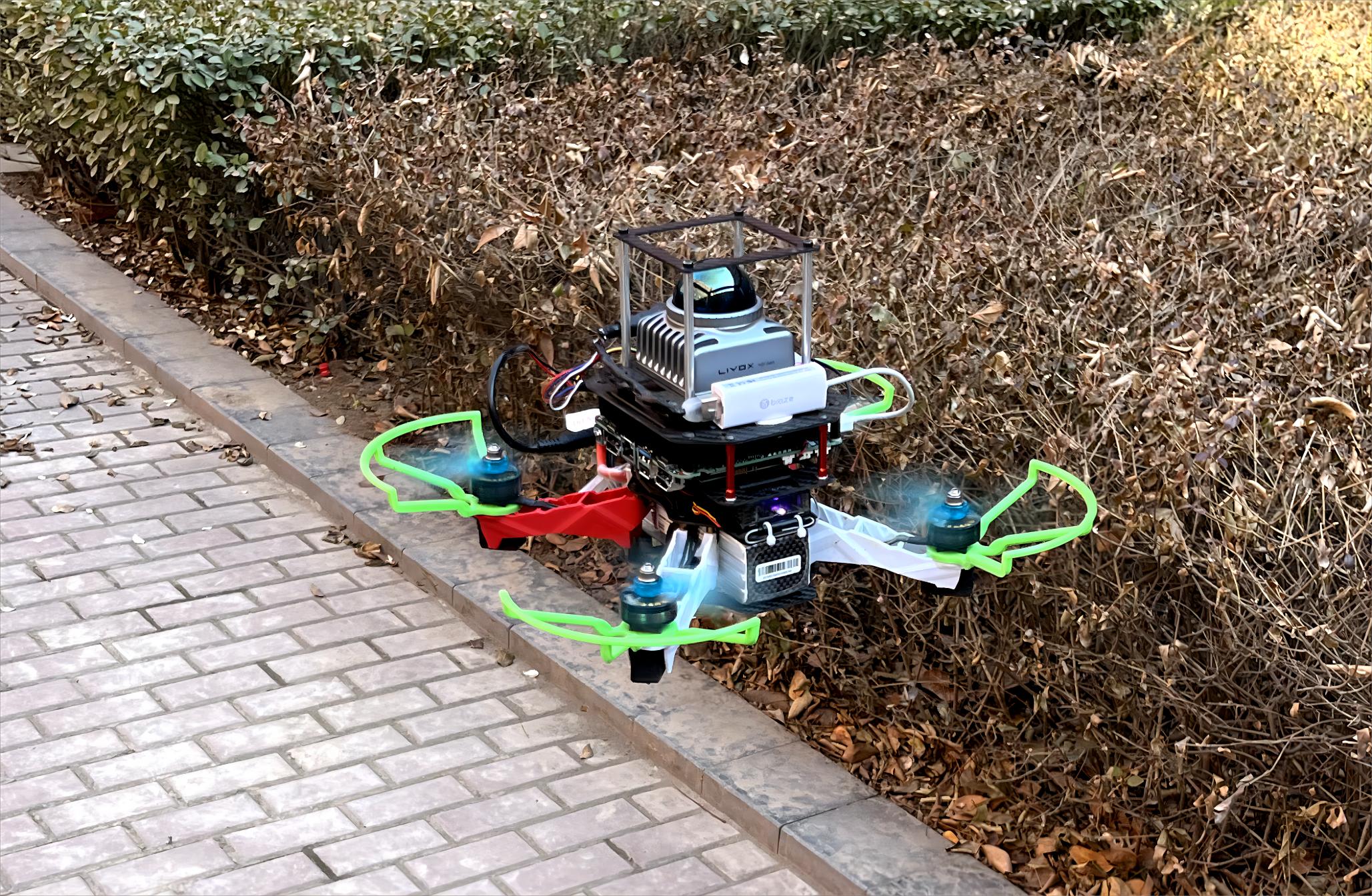}
\end{minipage}%
}%

\subfloat[]{
\centering
\begin{minipage}[htp]{0.85\linewidth}
\centering
\includegraphics[width=0.99\textwidth]{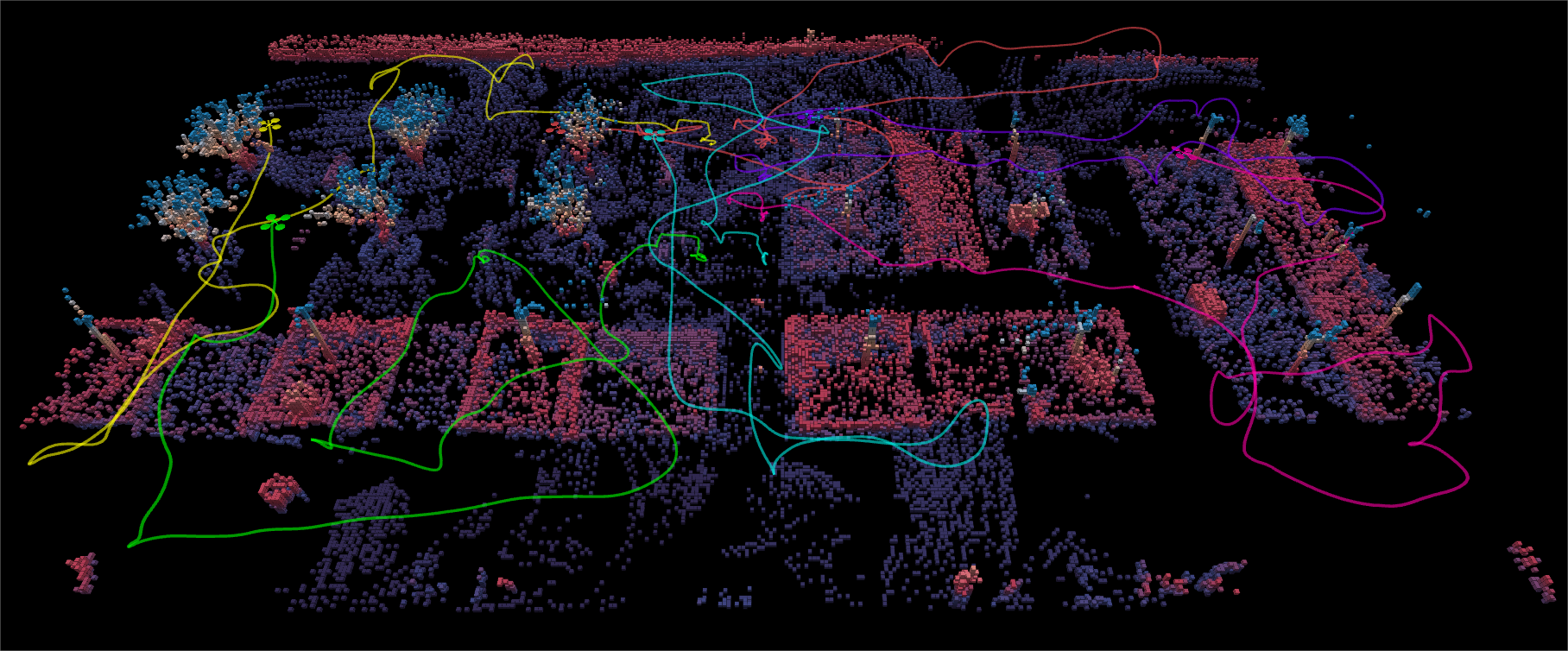}
\end{minipage}%
}%

\subfloat[]{
\centering
\begin{minipage}[htp]{0.85\linewidth}
\centering
\includegraphics[width=1.01\textwidth]{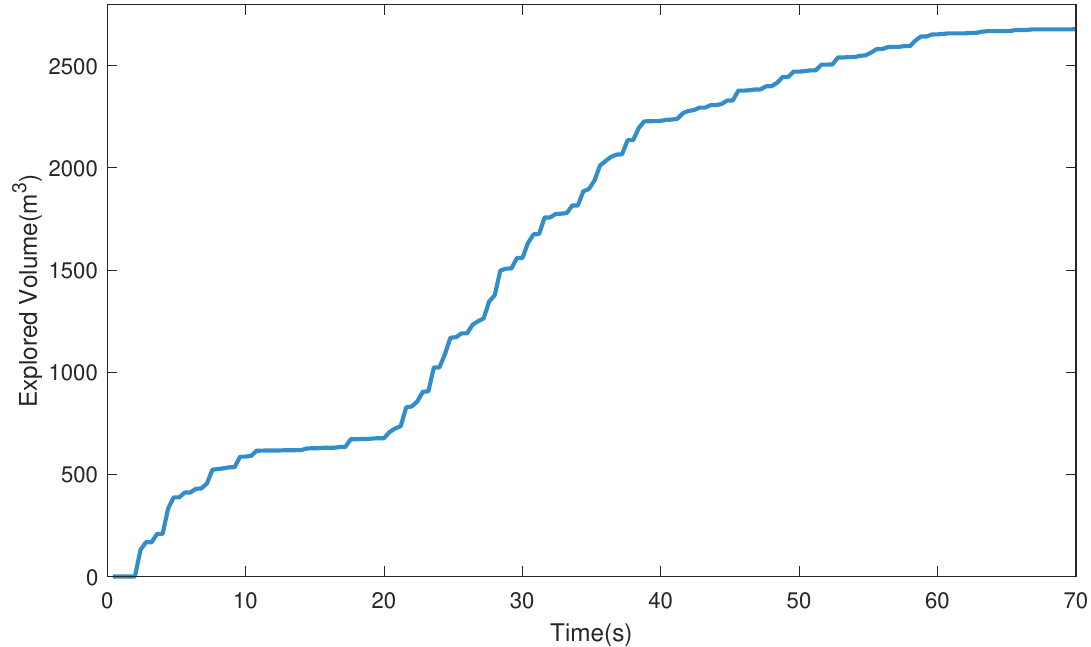}
\end{minipage}%
}%
\caption{(a) The outdoor environment. (b) The platform for outdoor exploration. (c) The mapping result. (d) The exploration progress of the proposed method in the outdoor experiment.}
\label{real}
\end{figure}
To further demonstrate the robustness and effectiveness of our method, a real-world experiment in the outdoor environment with 6 UAVs is conducted. Only on-board sensing and computation resources are used for each UAV. As shown in Fig.~\ref{real}(a), the outdoor scenario is with bushes and the size of it is $40\times24\times3\,\rm{m^3}$. As shown in Fig.~\ref{real}(c), the UAVs explored most of the explorable space within $67.2$s by the distributed trajectories. Fig.~\ref{real}(d) shows the exploration progress. In addition, the total communication volume of the multi-UAV system is only 14.47 MB.

In summary, both simulation and experimental results show that the proposed method achieved low communication traffic volume and fast exploration speed. More details of all experiments are available in the supplementary video. 

\section{CONCLUSION}
In this paper, we propose a hierarchical exploration method for fast multi-UAV autonomous exploration based on the fast-sharing multi-robot dynamic topological graph and graph Voronoi partition. It achieves higher exploration efficiency and lower communication volume compared with the state-of-the-art method. Moreover, the real-world experiment demonstrates that our method can be effectively applied in real-world 3-D multi-UAV exploration. 

One limitation of our method is the localization noise is not considered. In future work, we will take the localization of the multi-UAV system during the exploration into account.

\bibliographystyle{IEEEtran}
\bibliography{bibFile}

\end{document}